# Towards Constructing a Corpus for Studying the Effects of Treatments and Substances Reported in PubMed Abstracts

Evgeni Stefchov[1], Galia Angelova[2] and Preslav Nakov[3]

[1] Faculty of Mathematics and Informatics, Sofia University "St. Kliment Ohridski", 5 James Bourchier Blvd., 1164 Sofia, Bulgaria
[2] Institute for Information and Communication Technologies, Bulgarian Academy of Sciences, 25A Acad. G. Bonchev Str., 1113 Sofia, Bulgaria
[3] Qatar Computing Research Institute, HBKU, Doha, Qatar
`evgenistefchov@abv.bg, galia@lml.bas.bg, pnakov@qf.org.qa`

**Abstract.** We present the construction of an annotated corpus of PubMed abstracts reporting about positive, negative or neutral effects of treatments or substances. Our ultimate goal is to annotate one sentence (rationale) for each abstract and to use this resource as a training set for text classification of effects discussed in PubMed abstracts. Currently, the corpus consists of 750 abstracts. We describe the automatic processing that supports the corpus construction, the manual annotation activities and some features of the medical language in the abstracts selected for the annotated corpus. It turns out that recognizing the terminology and the abbreviations is key for determining the rationale sentence. The corpus will be applied to improve our classifier, which currently has accuracy of 78.80% achieved with normalization of the abstract terms based on UMLS concepts from specific semantic groups and an SVM with a linear kernel. Finally, we discuss some other possible applications of this corpus.

## 1 Introduction

PubMed is a large-scale database consisting of more than 28 million references and abstracts of biomedical publications. Using it as a knowledge discovery resource is tempting but challenging due to the highly specialized biomedical language with rich terminology and numerous abbreviations. Yet, PubMed abstracts are freely available, which is in contrast to other narratives in medical AI, e.g., clinical notes, which are inaccessible according to privacy regulations and requirements for in-domain knowledge. Other attractive features of PubMed as a text corpus include the fact that biomedical publications are written with grammatically correct sentences and more or less have a predefined discourse structure. This facilitates the application of Natural Language Processing (NLP) tools and supports various research tasks that rely on automatic identification of terminology and its patters of use.

Nowadays, text mining of PubMed abstracts is often the first step in the creation of annotated corpora as language resources supporting further information extraction tasks. Ambitious projects such as the development of databases with structured biomedical data and discovery of valuable associations between concepts depend on the correct automatic identification of the key semantic "carriers": *terms*, which also need normalization, *values* of essential attributes, and *sentences* that summarize the findings and the conclusions. The relevant literature shows that annotated corpora are usually prepared using a combination of automatic text processing and manual annotation and refinement.

In our task, we develop a manually annotated corpus in order to improve text classification. We want to identify one most important key sentence in every PubMed abstract that reports about negative, positive or neutral effects of some potential catalyst *X* on some medical condition *Y*. During the annotation process, we try to identify patterns about how these effects are verbalized in the texts and what the relation between the terms and abbreviations in the title and the rationale sentence is. Obviously, developing a large-scale corpus of this type is too expensive, slow and almost impossible, and thus we attempt to address this bottleneck primarily by pre-selecting PubMed abstracts that contain explicitly the key phrases *negative effect*, *positive effect* or *no effect* in their titles. In this way, we incorporate the authors' judgment about the abstract content and continue with the manual annotation of rationale sentences.

This paper is structured as follows. Section 2 lists some related work on corpus construction and text classification using PubMed abstracts; only few articles are mentioned among the numerous research papers dealing with PubMed texts. Section 3 explains specific aspects in the development of our manually annotated corpus using automatically selected, structured abstracts; various linguistic modalities encountered in these texts are considered as well as frequent patterns for the verbalization of negative, positive and neutral effects. Section 4 presents the application scenarios we intend to explore. Section 5 contains the conclusion and outlines plans for future work.

## 2 Related Work

We consider some research work dealing with corpus development. Information extraction was used on a corpus of PubMed abstracts to automatically identify and categorize biologically relevant entities and predicative relations by Zaremba & al. [1]. The relations include the following: Genes; Gene Products and their Roles; Gene Mutations and the resulting Phenotypes; and Organisms and their associated Pathogenicity. A total of 465 abstracts were collected and randomly split into a training set (327 abstracts) and a test set (138 abstracts). The training set was used for developing a lexicon and extraction rules. Specific annotation guidelines for entities and relations were elaborated by two molecular biologists and a computational linguist. Manual mark-up was performed by one biologist and reviewed by the other. Evaluations have shown very good accuracy, esp. given the relatively small training set: F-measure higher than 90% for entities (genes, operons, etc.) and over 70% for relations (gene/gene product to role, etc.).

Doğan & al. [2] present the disease corpus developed in the US National Center for Biotechnology Information (NCBI), which contains disease name and concept annotations in a collection of 793 PubMed abstracts. Each abstract was manually annotated by two annotators with disease mentions and their corresponding concepts in the medical vocabulary MeSH and the online catalog OMIM. Much attention was paid to achieving high inter-annotator agreement. The public release of this corpus contains 6,892 disease mentions mapped to 790 unique disease concepts. The corpus was used as a means for improving the recognition of disease names in real texts (so-called disease normalization, which is as a very difficult task). Using the corpus, three different disease normalization methods were compared, achieving an F-measure of 63.7%. The authors concluded that "these results show that the NCBI disease corpus has the potential to significantly improve the state-of-the-art in disease name recognition … by providing a high-quality gold standard thus enabling the development of machine-learning based approaches for such tasks".

Another dataset is PubMed 200k RCT, which contains about 200,000 abstracts of randomized controlled trials (RCT), and a total of 2.3 million sentences [3]. Each sentence is labelled with its role in the corresponding abstract: *background*, *objective*, *method*, *result*, or *conclusion*. Only PubMed articles with MeSH index D016449, corresponding to RCTs, were included in the dataset. In addition, these abstracts were explicitly structured into 3-9 sections and contained no sections labelled "None", "Unassigned", or "empty". When the labels of each section were originally given by the abstracts' authors, PubMed mapped them into a smaller set of standardized labels: "background", "objective", "methods", "results", "conclusions", "None", "Unassigned", or "" (an empty string). According to [3], more than half of the RCT abstracts in PubMed were unstructured. The expectation was that the public release of this dataset would accelerate the development of algorithms for sequential sentence classification; such tools in turn would facilitate sentence label prediction and hence, efficient browsing of literature because readers would be able to access selected sections only.

Concerning the amount of annotation samples necessary for a successful classifier training, Zaidan & al. [4] proposed to reduce the number of training examples needed, but to ask human annotators to provide hints to a machine learner by highlighting contextual "rationales" for each of their annotations. In this way, the annotator can identify features of the document that are particularly relevant and mark related portions of the example. Annotating rationales does not require the annotator to think about the classification feature space, nor even to know anything about it. So they demonstrated a method eliciting extra knowledge from naïve annotators, in the form of "rationales" for their annotations, which has significantly better performance than two strong baseline classifiers. It is interesting that their approach worked for positive and negative movie reviews across four annotators who had different rationale-marking styles.

Going deeper into the idea of using rationales (short and coherent pieces of input text, sufficient to provide motivations), we mention the work of Lei & al. [5] who have shown how to design a generator (for automatic generation of rationales) and an encoder (for predicting justification). The generated rationales are subsets of the words from the input sentences with two key properties: first, they represent short and coherent pieces of text (e.g., phrases) and, second, the selected words must alone suffice for prediction as a substitute of the original text. Here, we are interested in their manually annotated test corpus, which provided the evaluation of multi-aspect sentiment analysis on beer reviews. The evaluation was done on sentence-level annotations on around 1,000 beer reviews with multiple sentences describing five features: the appearance, smell (aroma), palate, taste and overall impression of a beer. The annotation of each sentence indicates what aspect this sentence covers. In addition to the written text, the reviewers provided the ratings for each aspect (originally on a scale from 0 to 5 stars) as well as an overall rating. However, we notice that the generator learns phrases as rationales, not full sentences. Moreover, there can be several phrases generated as rationales for the same input text and the same aspect. Here, we work with several rationales per abstract as well.

Text classification is a common approach for clinical text mining, as a single technique or as a component in more complex NLP environments. Usually, such systems are focused on specific diseases, drugs or facts. For instance, simple text classification helps to detect misdiagnoses of epilepsy West syndrome in pediatric hospital narratives [6]. A retrospective analysis was conducted on 27,524 patient records with diagnoses and a corpus of 3,744 records was constructed (for 144 patients as positive examples and 3,600 randomly selected ones as negative examples). Without any additional annotation, multinomial Naïve Bayes and Support Vector Machine (SVM) classifiers were run and compared, with SVM achieving precision 76.8%, recall 66.7% and F-measure of 71.4% when evaluated with 10-fold cross-validation. The authors note that "the use of domain knowledge is not a necessary requirement to achieve reasonable results" [6].

A more sophisticated example is a hybrid pipeline for heart disease risk factor identification that analyzes clinical texts and recognizes diseases, associated risk factors, associated medications, and the time they are presented [7]. This pipeline integrated rule-based processing and classification and achieved an F-measure of 92.68% at Track 2 of the 2014 i2b2 clinical NLP challenge. After preprocessing, this system extracted phrase-based, logical and discourse tags. Time attribute identification was interpreted as a classification task, which was solved using an SVM. The system generated candidate sentences, containing risk factors, and passed them to classifiers. The authors concluded that a possible improvement could be to generate less negative samples by limiting the candidates to those that contain medical concepts in UMLS.

Finally, we consider a more general approach for classifying patient portal messages [8]. This task is important because patient portals are increasingly adopted as communication means: patients express various needs and medical experts deliver recommendations as well as informal opinions. The main categories are informational, medical, logistical, social, and other communications, with subcategories including prescriptions, appointments, problems, tests, follow-up, contact information, and acknowledgements. Secure portal messages might contain more than one type of communication. The performance of the classifiers was evaluated using a gold corpus of 3,253 manually annotated portal messages.

# 3    Corpus Construction and Annotation

## 3.1    Selection of PubMed Abstracts

We want to find PubMed abstracts discussing the effect of some potential catalyst $X$ on some $Y$ under conditions $Z$. With the intension to collect the corpus faster, our first idea was to filter out PubMed abstracts that contain in their titles explicit statements about effect type: *negative*, *positive* and *neutral*. A sample title with this preferred structure is "*Positive effect of direct current on cytotoxicity of human lymphocytes*". However, even for these abstracts, the subsequent manual annotation showed that the justification of the decisions about the effect of catalysts $X$ on certain $Y$ is hard. There is a considerable amount of abstracts entitled "the effect of *X1* and *X2* on *Y*" or "the effect of *X* on *Y1* and *Y2*", for instance the title "*The mumps and rubella vaccination: no effect of feedback of vaccination scores in general practice*" and the title "*Positive effect of treatment with synthetic steroid hormone tibolon on intimal hyperplasia and restenosis after experimental endothelial injury of rabbit carotid artery*". The variety of patterns and potentially misleading constructions make the titles difficult to inter-pret and annotate:

- "*Negative effect of age, but not of latent cytomegalovirus infection on the antibody response to a novel Influenza vaccine strain in healthy adults*"
- "*Calcium influx inhibition: possible mechanism of the negative effect of tetrahydropalmatine on left ventricular pressure in isolated rat heart*"
- "*Positive effect of etidronate therapy is maintained after drug is terminated in patients using corticosteroids*".

Table 1 shows the amount of abstracts extracted initially from PubMed because their titles contained explicit description of the effect as *negative*, *positive* and *neutral*. In order to produce quickly a dataset where the annotation of the rationale will be maximally simple for naïve annotators, we removed all abstracts with "problematic" titles from the corpus. Table 2 contains the numbers of remaining abstracts after the removal of titles that contain "and", "or", "but", "review", "study", "meta-analysis", etc. Another problematic case are titles where the polarity is determined by a modifier preceding the statement about *negative*, *positive* and *no* effect, which is shown in a sample above "*absence of positive effect ...*" or, e.g., in the title "*Association of cystic fibrosis transmembrane-conductance regulator gene mutation with negative outcome of intracytoplasmic sperm injection pregnancy in cases of congenital bilateral absence of vas deferens*". In this way, we eventually ended up with 750 abstracts as shown in Table 3.

Note that by keeping only abstracts where the phrase *positive/negative/no effect* occurs in the beginning of the title, we avoid titles containing e.g. "*dominant negative effect*". The latter is ambiguous because "*dominant negative*" is a term in molecular biology, meaning a mutation whose gene product adversely affects the normal, wild-type gene product within the same cell. So, domain terminology is another important factor that needs to be taken into consideration when constructing the corpus. Further potentially ambiguous titles are treated properly, e.g., "*No effect of negative mood on the alcohol cue reactivity of in-patient alcoholics*" contains the word "*negative*", but it is considered in the neutral class because "*no effect*" is the first phrase of the title.

**Table 1.** Abstracts with *positive/negative/no effect* in the title.

| Pattern in the title | Effect of | Impact of | Influence of | Total |
|---|---|---|---|---|
| Positive | 242 | 133 | 52 | **427** |
| Negative | 247 | 238 | 50 | **535** |
| No | 782 | 84 | 127 | **993** |
| Total | 1,271 | 455 | 229 | **1,955** |

**Table 2.** Abstracts without *and/or/but/review/study/meta analysis*, etc. in the title.

| Pattern in the title | Effect of | Impact of | Influence of | Total |
|---|---|---|---|---|
| Positive | 135 | 74 | 27 | **236** |
| Negative | 171 | 140 | 31 | **342** |
| No | 406 | 40 | 82 | **528** |
| Total | 712 | 254 | 140 | **1,106** |

**Table 3.** Abstracts with *positive/negative/no effect* at the beginning of the title.

| Pattern in the title | Effect of | Impact of | Influence of | Total |
|---|---|---|---|---|
| Positive | 86 | 57 | 19 | **162** |
| Negative | 63 | 73 | 18 | **154** |
| No | 341 | 30 | 63 | **434** |
| Total | 490 | 160 | 100 | **750** |

### 3.2 Annotating the Rationale

In Section 2, we listed approaches to classification based on rationales (one sentence or phrases [4, 5] that represent short and coherent pieces of text that must suffice to express the polarity of the effect presented in the text). However, both approaches [4, 5] deal with product reviews: [4] works with one rationale while [5] (dealing with multi-aspect classification) allows multiple rationales per input text. Here, we apply the same idea to biomedical abstracts, which are scientific publications, i.e., their primary aim is not to assess some specific product (movie, bier, etc.). Thus, the rationale is often expressed in several sentences listing results of tests with target patients groups. We found that the summarizing concluding sentences usually generalize the findings and provide some argumentation about the polarity of the effect, but the latter is often expressed by domain-specific verbalizations. One sample abstract is shown in Figure 1. In contrast to product reviews that contain numeric scores in addition to the text, the variability of nuances in PubMed abstracts is much higher and we need biomedical expertise to decide about the "most important, most convincing" sentence or phrases that must suffice to express alone the polarity of the effect.

---

Title[1]: **Negative effect of aging on psychosocial functioning of adults with congenital heart disease**

ABSTRACT:

BACKGROUND: Improvements in life expectancy among adults with congenital heart disease (ACHD) provide them with unique challenges throughout their lives and age-related psychosocial tasks in this group might differ from those of healthy counterparts. This study aimed to clarify age-related differences in psychosocial functioning in ACHD patients and determine the factors influencing anxiety and depression.

METHODS AND RESULTS:A total of 133 ACHD patients (aged 20-46) and 117 reference participants (aged 20-43) were divided in 2 age groups (20 s and 30 s/40 s) and completed the Hospital Anxiety and Depression Scale, Independent-Consciousness Scale, and Problem-Solving Inventory. Only ACHD patients completed an illness perception inventory. _ACHD patients over 30 showed a significantly greater percentage of probable anxiety cases than those in their 20 s and the reference group_. Moreover, ACHD patients over 30 who had lower dependence on parents and friends, registered higher independence and problem-solving ability than those in their 20 s, whereas this element did not vary with age in the reference participants. _Furthermore, ACHD patients may develop an increasingly negative perception of their illness as they age_. The factors influencing anxiety and depression in patients were aging, independence, problem-solving ability, and NYHA functional class.

CONCLUSIONS: Although healthy people are psychosocially stable after their 20 s, _ACHD patients experience major differences and face unique challenges even after entering adulthood._

**Fig. 1**. Phrases with abbreviations supporting the polarity in Results and Conclusions.

In Figure 1, the rationales (short and coherent pieces of text) are underlined. We see that the automatic recognition and processing of abbreviations is very important because often the rationales contain abbreviated versions of the basic terms that are defined in the particular abstract.

We should also note the complexity of assigning a category to each biomedical abstract. Naïve annotators, who are not medical professionals, have limited capacity to comprehend and to interpret the abstract content, and thus they tend to make decisions based on general lexica, but they might be confused by the sentiment expressed in the text, as shown in the abstract on Figure 2. The underlined sentences convince the ordinary reader that no negative effect has been observed, and in this way the title is a bit misleading and the abstract should be annotated as reporting about a "_neutral effect_". Only experts might judge whether the phrases in italic (_pleural thickening_, _small foci of collagen_) are dangerous deviations in the "formation of 8-hydroxydeoxyguanosine in DNA", which is the subject according to the title.

---

[1] https://www.ncbi.nlm.nih.gov/pubmed/25391256

The Japanese authors of this abstract perhaps intended title to be "*On the negative effect of ...*". This example shows that all texts included in the initially selected corpus of PubMed abstracts need careful manual assessment and refinement. Only few such abstracts were found and removed so far, all of them authored by non-native speakers.

---

**Title[2]: Negative effect of long-term inhalation of toner on formation of 8-hydroxydeoxyguanosine in DNA in the lungs of rats in vivo**

**Abstract**: We assessed the effects of long-term inhalation of toner on the pathological changes and formation of 8-hydroxydeoxyguanosine (8-OH-Gua) in DNA in a rat model. Female Wistar rats (10 wk old) were divided evenly into a high concentration exposure group (H: 15.2 mg/m(3)), a low concentration exposure group (L: 5.5 mg/m(3)), and a control group. The mass median aerodynamic diameter of the toner was 4.5 microm. The rats were sacrificed at the termination of a 1-yr or 2-yr inhalation period. Pathological examination was performed on the left lung, and the level of 8-OH-Gua in DNA from the right lung was measured using a high-performance liquid chromatography (HPLC) column. <u>The pathological findings showed that lung cancer was not observed in any of the exposed or control groups</u>, though *pleural thickening and small foci of collagen were observed in toner-exposed rat lungs*. Inhalation of the toner for 1 and even 2 yr did not induce the formation of 8-OH-Gua in DNA in rat lungs. <u>These data suggest that long-term inhalation of toner may not induce lung tumors.</u>

---

**Fig. 2**. An abstract that might look neutral to naïve annotators.

### 3.3    Typical Expressions and Frequent Patters

The typical titles of the PubMed abstracts we consider have the following structure:

[ the|a ] negative effect of $X$ on|in|for $Y$

[ the|a ] positive effect of $X$ on|in|for $Y$

[ the|a ] no|neutral effect of $X$ on|in|for $Y$

The grammatical correctness of the PubMed titles enables easier automatic recognition of terms and their abbreviations. While progressing with the annotation, we discover also typical patterns of phrases that signal the effect polarity, for instance:

- In the "*no effect*" abstracts: "*Overall results disclosed no significant effect of this drug on ...*", "*no significant difference was observed ...*", "*there were no notable changes*", "*X is unlikely to cause clinically significant interactions*", "*X does not have any detrimental effect on the longevity and clinical outcomes*"

- In the "*positive effect*" abstracts: "*X can facilitate the normalization of Y*", "*X may be effective in leading to Y*", "*X is likely to be worthwhile and unlikely to be harmful for Y*", "*X leads to improved quality of life over 96 weeks*", "*X was found to increase in yeast cells*", "*Patients exposed to the integrated care model exhibited significantly fewer depressive symptoms*".

---

[2] https://www.ncbi.nlm.nih.gov/pubmed/16195210

- In the "*negative effect*" abstracts: "*the prolonged use of X may have a negative effect on Y*", "*X has a relationship not only with Y, but also with a poor outcome*", "*X is a significant prognostic factor for low overall survival*", "*Development of X is associated with worse long-term outcomes*".

When selecting the rationale sentences, we choose among the candidates those that contain the terminology mentioned in the abstract's title.

## 4    Current Classification Experiments

In previous work, we studied the classification of PubMed abstracts without using any manually annotated corpus [9]. Various experiments were performed with text preprocessing options, semantic indexing using UMLS and MetaMap, and classification algorithms like Bernoulli/Multinomial Naïve Bayes classifiers and Linear/RBF SVM.

Abstracts are represented as bags of words, but they can have sections (segments) with subtitles: e.g., Background, Results, Conclusions, etc. This structure is either provided by PubMed or we induce it automatically. Segment names are not fixed and, in addition to discovering the segments, we also map their names to a predefined set – Background and Objectives, Methods, Results, Conclusions and Others. The label "Others" is assigned to text fragments without any explicitly assigned subtitle.

Different variants are proposed for preprocessing the collected abstracts, which include the use of concepts from semantic networks such as UMLS and MeSH. The semantic indexing is performed by MetaMap [10, 11] – the conceptual search engine of UMLS that provides indexing of terms from the given text by user-selected vocabularies included in UMLS. In one of the experiments, UMLS concepts were limited by semantic groups to a predefined set.

One group of tests explicates the role of the different abstract sections. We give different inputs to the classifiers – the full abstract, only the Conclusions or the Results section, or both Results and Conclusions. We specifically chose the Results and the Conclusions sections because the usefulness of the effect should be most explicit there. A second cycle of experiments used the results of the initial ones, e.g., if the best results for linear SVM are achieved using the full abstract, then the full abstract will be considered in further experiments when classifying with Linear SVM. The second type of experiments determines the importance of domain knowledge by importing names of concepts found in UMLS and MeSH, where the concepts might be limited by semantic groups. In these experiment, concepts from UMLS or the MeSH hierarchy will be added to or replace the corresponding terms in the sentences where they occur.

We also included the title in these experiments. We tried to define the sentence from the Results or the Conclusions section that has the largest word overlap (without stop words and punctuation) with the title and to classify the given example only based on this 'best sentence'. Word overlap is defined as the overlap of raw words from the original text or as overlap of the found UMLS/MeSH concepts. Then we use the concepts from UMLS and MeSH that are found in the title; the idea is to normalize the abstract by replacing each occurrence of the UMLS/MeSH concept from the title with the tags $X1, X2,.., Xn$ where $n$ is the number of concepts found in the title.

We define two baselines to compare with the results of our system. The first one is a 'majority classifier' that always predicts the most frequent class in our dataset. Its accuracy is 57.87%. The second baseline predicts the class of the example by searching for the same signaling phrases as in the title, but this time in the article's abstract and choosing the most frequent class as per these phrases, breaking ties using the first baseline; the accuracy for this second baseline is 61.60%.

The best results for the experiment that defines the role of the different parts of text is achieved by using the full abstract and linear or RBF-based kernel SVM, with accuracy of 76.27% and 75.47%, respectively. When we use the full abstract and we limit the concepts from UMLS by semantic type, we achieve slightly better accuracy of 76.80%. If the classification is done only by using the 'best sentence', the accuracy decreases to 74.50%. This can be explained by the fact that although one sentence has the largest word overlap with the title, this does not necessarily mean that it will most clearly determine the polarity of the abstract. Our best accuracy of 78.80% is achieved with UMLS normalization of the abstract, limited by semantic type, and using a linear SVM classifier. We are not aware of other experiments that consider in such details the textual structure of the abstracts, the weight of the title' words and the role of semantic terminology processing. We expect that training on the manually annotated rationales, the accuracy will improve further.

## 5 Conclusion and Future Work

We have presented our on-going efforts towards the construction of a corpus containing PubMed abstracts annotated with rationales. Previous work has shown that using rationales provides substantial improvements for classifying product reviews [4, 5], and now we want to test this idea on PubMed abstracts. Several questions remain open at this stage of our work, e.g., whether it is better to mark one rationale per abstract or we can annotate several ones. Another open question is whether a corpus of 750 abstracts will allow us to see significant improvements on the classification task given the large medical vocabulary and the comprehensive PubMed archive.

In future work, we plan experiments with various techniques for abstract classification, e.g., using clustering [12] and deep learning [5]. We consider very important the automatic processing of abstract terminology with special focus on the abbreviations, to enable term normalization and reference to the abbreviated forms. This will help to connect the terms in the abstract titles and the abbreviations occurring in the rationale phrases and sentences. In this way, we could incorporate rule-based analysis in the preprocessing phase. Another promising research direction is on using summarization [13] techniques to find the most important part of the abstract, and then performing classification based on this abstract, or using attention mechanism to focus the model on sentences in the abstract that are most similar to the title; the latter is typically done in an unsupervised way, but we could use our manual rationales to train a supervised attention mechanism [14]. We also plan to experiment using various word representations, which have been shown useful for biomedical text classification [15].

**Acknowledgements.** The authors are grateful to the anonymous reviewers for their valuable comments.